\documentclass[11pt]{article}
\usepackage[margin=1in]{geometry}
\usepackage{array}
\usepackage{booktabs}
\usepackage{etoolbox}
\usepackage{float}
\usepackage{graphicx}
\usepackage{microtype}
\usepackage[font=small,labelfont=bf]{caption}
\usepackage[section]{placeins}
\usepackage[title]{appendix}
\usepackage{xurl}
\usepackage[hidelinks]{hyperref}
\useOriginalUrlSetting
\usepackage{amsmath}
\usepackage{longtable}

\hypersetup{
  pdftitle={Same Model, Different Harness: Different Coding-Agent Results},
  pdfauthor={Sydney Lewis}
}

\makeatletter
\@ifpackagelater{longtable}{2025/12/22}{}{%
  \patchcmd{\LT@output}
    {\copy\LT@foot\vss}
    {\copy\LT@foot\vskip\z@\@plus 1fil\@minus\normalbaselineskip}
    {}{\PackageError{paper}{First longtable finite-shrink patch failed}{}}
  \patchcmd{\LT@output}
    {\copy\LT@foot\vss}
    {\copy\LT@foot\vskip\z@\@plus 1fil\@minus\normalbaselineskip}
    {}{\PackageError{paper}{Second longtable finite-shrink patch failed}{}}
}
\makeatother
\newcommand{\ControlCredKappa}{0.66}
\newcommand{\ControlCredN}{500}

\newcommand{\ControlCredSamePct}{87.6}

\newcommand{\DialCohortN}{15}

\newcommand{\FBN}{183}

\newcommand{\ProCN}{316}

\newcommand{\VCohortN}{169}

\newcommand{\VSevC}{43}

\newcommand{\VSevFtpMeanC}{28}
\newcommand{\VSevFtpMeanT}{49}

\newcommand{\VSevT}{72}

\newcommand{\QFullFBPressCommonDenomFtpC}{10.53}
\newcommand{\QFullFBPressCommonDenomFtpCHigher}{9}
\newcommand{\QFullFBPressCommonDenomFtpT}{19.38}
\newcommand{\QFullFBPressCommonDenomFtpTHigher}{47}

\newcommand{\QFullFBPressFtpC}{10.5}

\newcommand{\QFullFBPressFtpT}{19.6}

\newcommand{\QFullFBPressN}{183}
\newcommand{\QFullFBPressResC}{2}

\newcommand{\QFullFBPressResT}{3}

\newcommand{\QFullFBWideFtpC}{23.9}

\newcommand{\QFullFBWideFtpP}{0.00022}

\newcommand{\QFullFBWideFtpT}{30.7}

\newcommand{\QFullFBWideN}{183}
\newcommand{\QFullFBWideResC}{5}

\newcommand{\QFullFBWideResT}{5}

\newcommand{\QFullVWideCompressionN}{153}
\newcommand{\QFullVWideCompressionReduction}{7.2}

\newcommand{\QFullVWideServedN}{154}

\newcommand{\QFullVWideTurnsIncrease}{1.4}

\newcommand{\QGradN}{169}

\newcommand{\QGradWideFtpDiff}{-0.3}
\newcommand{\QGradWideFtpDiffCIHi}{+3.9}
\newcommand{\QGradWideFtpDiffCILo}{-4.5}

\newcommand{\PaperRepository}{\url{https://github.com/sydches/yuj}}

\title{Same Model, Different Harness: Different Coding-Agent Results}
\author{Sydney Lewis}
\date{August 2026}

\begin{document}
\maketitle

\begin{abstract}
A coding agent combines a model with a harness, which decides what the model
sees, which tools it can use, and how the work continues. We ask whether
changing the harness changes the result when the model and task stay fixed. We
compare two configurations of the same harness on three coding benchmarks. The
control supplies the full conversation in time order, while the treatment keeps
the same record but mechanically shortens older tool results as the context
fills and responds to repeated or stalled work.
Under tight context, the treatment raises mean per-task fail-to-pass fraction
(F2PF) in all three pressure comparisons and increases complete solutions on
SWE-bench Verified and SWE-bench Pro. The tight-window Verified comparison uses
\VCohortN{} tasks, a 20{,}480-token window, and a fixed 480-second attempt
endpoint; on this cohort, treatment raises mean per-task F2PF from
\VSevFtpMeanC\% to \VSevFtpMeanT\% and complete solutions from \VSevC{} to
\VSevT{}. Without model-specific retuning, the same frozen treatment also
raises both endpoints on the same cohort for three additional models with
different designs. In the wide-window Qwen3.6 comparisons, observed arm
outcomes are close on Verified and Pro, while FeatureBench retains a higher
mean per-task F2PF under treatment. On the wide-window Verified cohort,
treatment also serves fewer prompt tokens per turn. Because changing the
harness changed what unchanged model weights could accomplish, coding-agent
evaluations should treat the model and harness together as the tested solver.
\end{abstract}

\section{Introduction}
A coding agent begins with an issue and a repository. It searches for
the relevant code, opens files, runs tests, makes edits, and tests
again. Each step leaves more text behind: commands, outputs, errors,
and the model's own replies.
On a large codebase, that history can become part of the problem. One
investigation may produce full files, broad search results, failed
commands, and long test logs, all of which compete for the same finite
context window.

Changing the model or its context window is one lever, while changing how the
harness uses the available working space is another. Before every new step, a
harness decides which text the model will see, supplies the tools, and decides
whether the run may continue. These choices may look like background plumbing,
but they are part of the solver.
Context pressure makes the contrast between harness configurations measurable,
although we study the harness, not context pressure itself. The control
configuration presents the history as one growing transcript in time order and
stops the run when the transcript fills the context window, even if the patch
is incomplete. Must the model's working view grow in the same way as the
permanent record of its work?

We built Yuj~\cite{yuj}, the coding-agent harness evaluated here, to test
another path. Yuj keeps a durable record of the run but can show the model a
smaller, changing view. It also watches for simple patterns such as repeated
failing commands or repeated reading without an edit. When such a pattern
appears, the harness applies a fixed response and watches what happens next.
The harness thus observes, acts, and checks again. We call this a
\emph{closed-loop harness}.
These mechanisms could help or harm. A smaller view can make room for the next
step but also hide a useful detail, while a response to repeated work can break
a loop but also interrupt useful repetition. We therefore ask whether the same
model repairs more required behavior and finishes more tasks, or takes a
different path without improving either endpoint.

We tested the treatment against the same harness's control
configuration. Within each pair, both arms used the same model weights, tasks,
context capacity, and run protocol. We measured complete solutions and the
share of required failing tests that each patch made pass.

This work grew out of direct experience building and operating coding-agent
harnesses. The references identify the specific models, benchmarks, software,
and reports that the study uses; they do not define an intellectual lineage for
the work. We claim no historical priority over earlier or independently
developed work and no novelty for the individual treatment mechanisms. Our
contribution is the treatment package as tested and the paired evidence that
changing the harness changes what the same model accomplishes on the same tasks.

\section{From a growing transcript to a closed loop}
\label{sec:instrument}

Each paired comparison used frozen control and treatment conditions. Under
control, the model received the full chronological conversation at each call,
and the run stopped when that conversation no longer fit in the context window.
Here, ``full'' means the full model-visible conversation after the ordinary
tool-result cleanup shared by both arms, not raw terminal bytes. Under
treatment, the conversation remained chronological, but fixed rules shortened
older tool-result bodies as the window filled, responded mechanically to
detected stall patterns, and guarded against a small set of predictable command
problems. Treatment did not replace the conversation with semi-structured
fields, a model summary, or semantic state.

\begin{figure}[H]
\centering
\includegraphics[width=\linewidth]{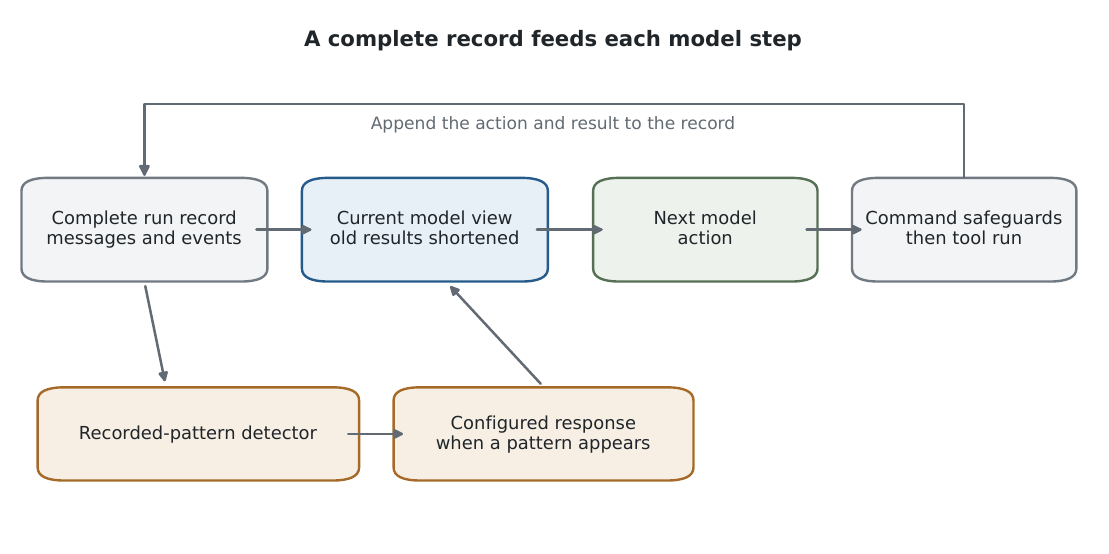}
\caption{The closed loop. The harness retains the in-memory conversation but
rebuilds the model's current view for each step. Recorded execution patterns
can change a later step. Separate files preserve the complete run record.}
\label{fig:closed-loop}
\end{figure}

Figure~\ref{fig:closed-loop} shows how these mechanisms connect.

\FloatBarrier

\subsection{The model's working view}

At each step, the model receives a block of text that includes the task, the
conversation so far, commands, and tool results. This block is the model's
\emph{context}, and its \emph{context window} fixes how much it can read at
once.
The window is working space for the model's next decision, not the permanent
archive of everything that has happened. On a long task, old tool results and
new evidence compete for this same limited space.
Control instead builds this context as one transcript in time order: the
harness adds new work to the end but keeps old searches, test logs, and errors
in front of the model. The run ends when that transcript reaches the window
limit, so a large early result can consume space needed later.

\subsection{Keep the record, change the view}
\label{sec:instrument:store}

The treatment separates the complete in-memory conversation from the model's
working view. During a run, it holds the conversation and tool results collected
so far in memory and uses them to build the next view. That view can shrink
while the in-memory conversation remains complete, and separate trace and
transcript files preserve the full run record.

New tool results enter the view in full, while fixed rules shorten older tool
results in stages as the window fills. The task, the model's own replies, and
other conversation messages remain unchanged. The mechanical policy shortens
displayed tool results directly, without a model summary, field extraction, or
reconstruction from trace or state files. It leaves the in-memory conversation
complete, constructs the view deterministically, makes each change directly
auditable, and requires no additional model call.

The study does not compare the mechanical rule with model-written summaries or
retrieval-based views. We leave that comparison for future work.

\begin{figure}[H]
\centering
\includegraphics[width=0.8\linewidth]{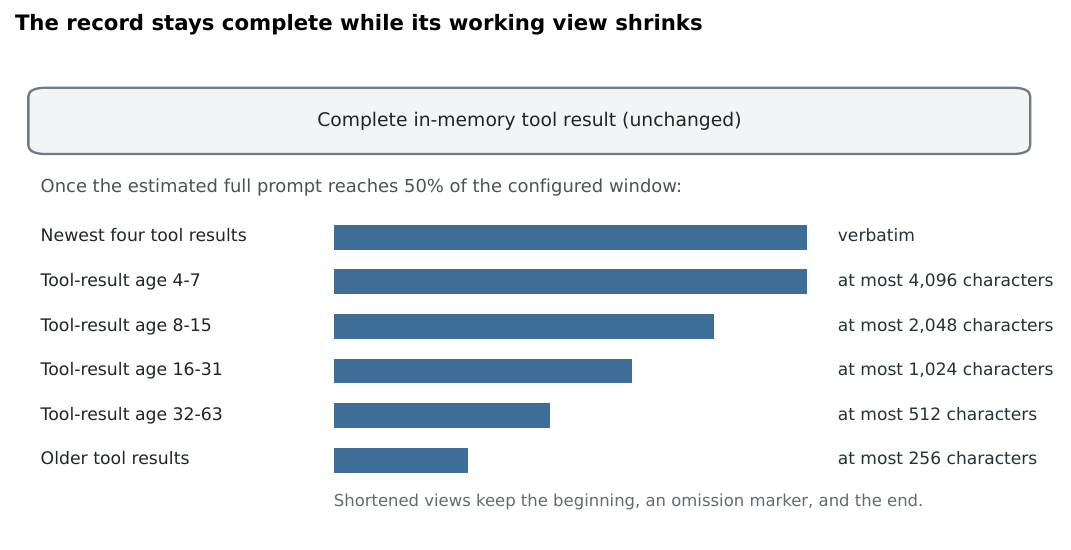}
\caption{The treatment's working-view rule. The treatment starts shortening
only after the estimated full prompt reaches half the configured window. The
newest four tool results remain full. Older results keep their beginning and
end within the printed caps, while the in-memory result remains complete.
Capped bar lengths use a base-two logarithmic scale. The printed values are
exact.}
\label{fig:view-decay}
\end{figure}

Figure~\ref{fig:view-decay} shows the exact stages.

Here, age is the number of newer tool results that have arrived. Once context
pressure activates the rule, the same age-and-length policy applies to every
tool result, while the detector below handles repetition separately. A
shortened result keeps its beginning and end around a plain omission marker.
The policy takes its name from this schedule: as age doubles across the
displayed tiers, the character cap halves. The half-life policy can be
configured along several axes that set the activation point, the number of
newest tool results kept in full, and the character cap for each age tier. We
froze these values within each paired comparison.

Shortening frees room for another search, edit, or test. The detector below
handles the separate problem of repeated or stalled work.

\subsection{Notice when the work stalls}

A run can have room left and still make no progress when the model repeats a
failing command, receives the same error several times, or rereads the same
material without making an edit. The harness therefore uses a simple,
consistent checker called the \emph{detector}.

The detector does not read or reformat the transcript. It looks at fixed facts
already in the execution record. For example, if the model repeats a command
and gets the same failure again, the record contains both the repeated command
and the repeated result, so the detector identifies that pattern. The model
still decides what code to change. We fixed the detector rules before the runs,
so the same recorded pattern maps to the same decision. The check uses no model
call or model tokens, and its rules and thresholds are part of the harness and
treatment configuration.

\FloatBarrier
\subsection{Act, then look again}

When the detector finds a pattern, the harness responds. In the
repeated-failure example, the response points out that the same attempt
produced the same failure and tells the model to take a different action toward
the task. The model then continues working, its next action joins the record,
and the detector keeps watching. We call the delivered response the
\emph{intervention}.

This creates a loop: observe the run, respond, then observe what happens next.
The treatment also handles a few predictable command problems by putting a test
command into the expected form, stopping a forbidden command before it runs,
and preventing oversized setup output from taking over the context. These fixed
safeguards keep avoidable command trouble from wasting the next model step.

The treatment condition combines the working-view rule, the detector and its
interventions, and the command safeguards.

% Experimental design.

\section{Comparing treatment with control}
\label{sec:design}

\subsection{One paired change}

We run every selected task in a primary pressure comparison once under control
and once under the treatment defined in Section~\ref{sec:instrument}. Within
each pair, both arms use the same task, model weights, context capacity, tool
interface, serving setup, evaluator, harness code, and fixed run and scoring
protocol. The control and treatment differ only in one checked-in configuration
package, so pairing estimates the total effect of assigning the treatment
condition as run. It does not identify the separate effect of any one treatment
rule. We measure complete resolution and the fail-to-pass fraction (F2PF),
which Section~\ref{sec:design:outcomes} defines in full.

We set the half-life operating point during earlier harness development, which
included operational runs from the same benchmark families. We froze the
operating point before the reported paired campaigns but did not select it
through a task-disjoint hyperparameter study. The comparisons therefore
evaluate the package as developed, not a held-out tuning procedure.

\begin{figure}[H]
\centering
\includegraphics[width=\linewidth]{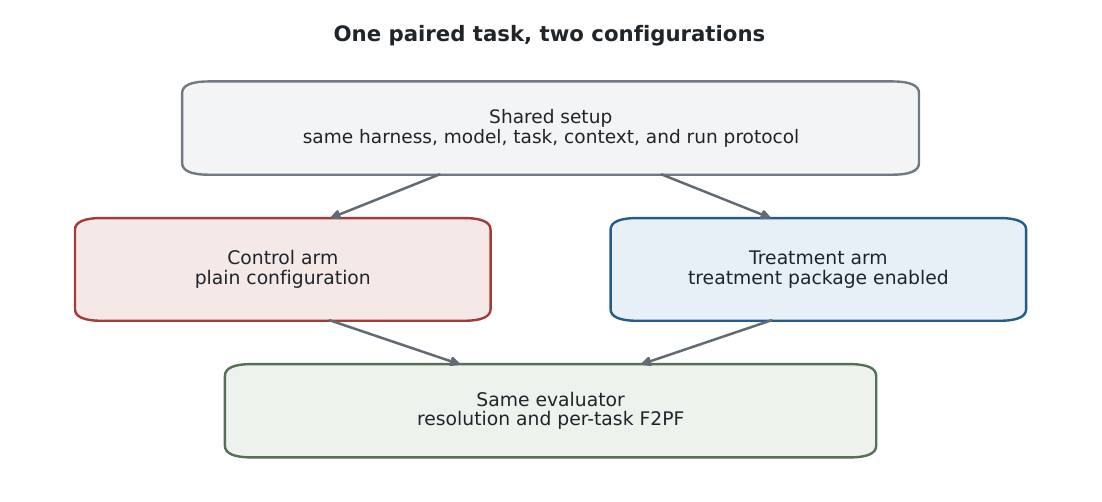}
\caption{The primary paired experiment. Both arms use the same model-facing
harness code, model, tasks, context capacity, run protocol, serving
setup, and evaluator. Configuration layers enable the treatment condition on
the treatment arm.}
\label{fig:paired-comparison}
\end{figure}

Figure~\ref{fig:paired-comparison} summarizes the paired design.

\FloatBarrier

The two runs unfold independently from the same starting conditions. In each
arm, the harness decides what to show the model and what to do next as the work
develops.

Before testing treatment, we checked that untreated Yuj provided a credible
control. We ran all \ControlCredN{} SWE-bench Verified tasks with GPT-5.5 through
Codex CLI under untreated Yuj and mini-SWE-agent v2.2.8~\cite{minisweagent},
keeping the model, tasks, and strict execution substrate fixed. The two
harnesses agreed on \ControlCredSamePct\% of task outcomes (Cohen's
kappa = $\ControlCredKappa$). This agreement supported using untreated Yuj as
the control for a within-Yuj comparison that changed only the configuration
package.

\subsection{Models, benchmarks, and task sets}

The primary comparisons use Qwen3.6-35B-A3B~\cite{qwen36_35b_a3b}, served with
\texttt{llama-server} from the \texttt{llama.cpp} project~\cite{llamacpp} and
four-bit Q4\_K\_XL weights. We vary the window size to create controlled
context pressure.
We test on SWE-bench Verified~\cite{swebench,swebench_verified},
SWE-bench Pro~\cite{swebench_pro}, and
FeatureBench~\cite{featurebench}. Each benchmark uses its own evaluator, but
each paired comparison uses the same task set in both arms.

Each benchmark also uses its own task-preparation protocol: locally sealed
images for Verified, the benchmark mask for FeatureBench, and benchmark-supplied
per-task images for Pro. In every comparison, both arms begin from the same
prepared task state, run without network access, and cannot access gold patches
or scorer-only inputs during model steps. Appendix~\ref{app:task-isolation}
gives the benchmark-specific controls.

Context pressure varies by task because some tasks require more reading before
the first source change. We call that amount the task's \emph{reading demand}.
Verified uses the same \VCohortN{}-task cohort at every context size and on
every model, while Pro uses a \ProCN{}-task demand-screened cohort and
FeatureBench uses all \FBN{} canonical tasks runnable in our evaluation setup.
Appendix~Table~\ref{tab:cohort-rules} gives the selection rules.

The Verified comparisons use the current frozen treatment package. The Pro and
FeatureBench pressure comparisons use the earlier complete version frozen for
those campaigns. Their 262{,}144-token rows come from earlier campaigns and are
not exact replications of the current package. The FeatureBench wide-window row
pairs an earlier treatment with a later corrected control.
Appendix~\ref{app:experimental-custody} records the run-era differences and
code-identity coverage.

\subsection{One treatment run, step by step}

\begin{figure}[H]
\centering
\includegraphics[width=\linewidth]{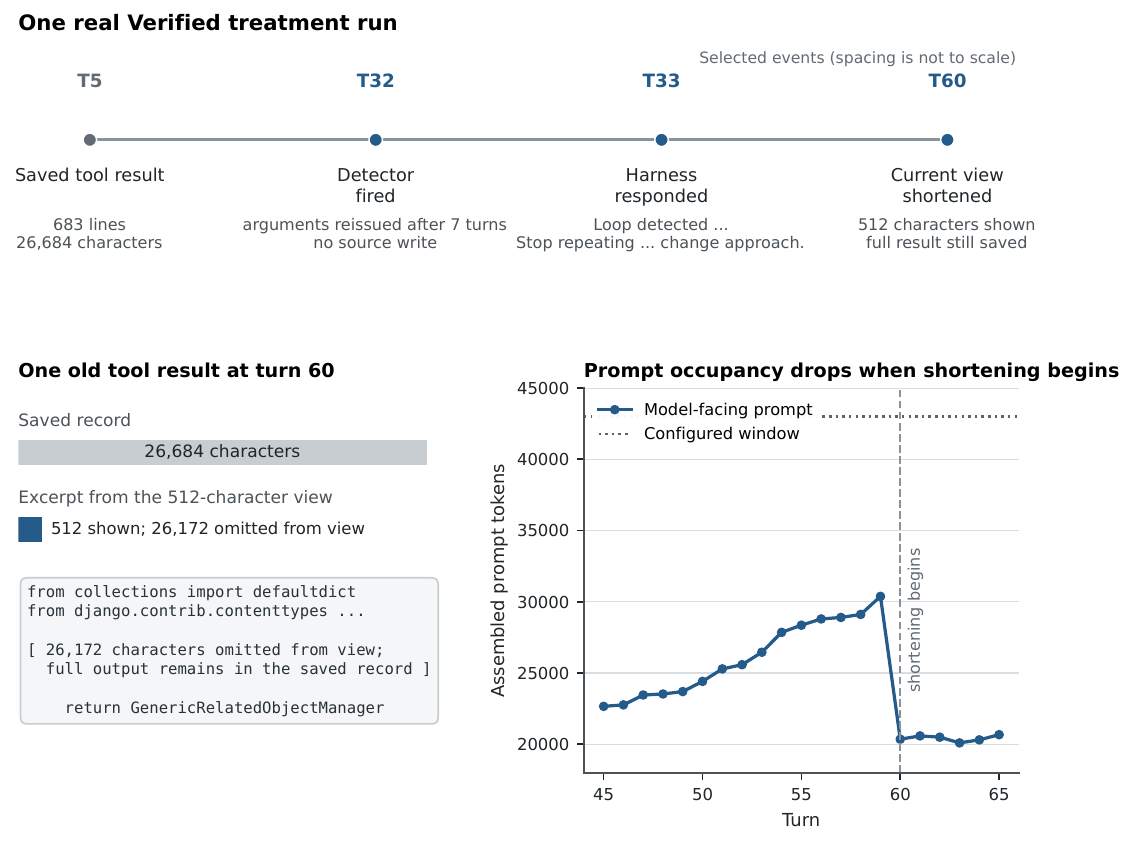}
\caption{A real Verified treatment trace. The lower-left panel shows an
excerpt from the exact 512-character view of a 26{,}684-character tool
result. The full result remained saved. The lower-right panel shows
assembled prompt tokens before and after shortening began at turn 60.
The top row reports the recorded detector fact and the reminder that
followed.}
\label{fig:worked-trace}
\end{figure}

Figure~\ref{fig:worked-trace} shows the treatment at work in one saved Verified
run for task \texttt{django\_\_django-11211}.

In this run, the treatment started shortening when the growing view needed
room. Table~\ref{tab:shortening-activity} shows how often shortening operated
across the primary treatment populations, when it began, and how much text it
typically removed. The summaries give every active run equal weight, and
Appendix~\ref{app:experimental-custody} records the calculation.

\begin{table}[H]
\centering
\caption{Shortening activity in each primary treatment population. The \emph{Active runs} column counts sessions with at least one shortening event. The \emph{First turn} column marks when shortening first removes characters. The \emph{Typical reduction} column reports the median share of characters removed. Brackets give the middle half of active runs.}
\label{tab:shortening-activity}
\small
\begin{tabular*}{\linewidth}{@{\extracolsep{\fill}}lrrr@{}}
\toprule
Cell (context window) & Active runs & First turn & Typical reduction \\
\midrule
Verified 43{,}008 & 107/169 (63.3\%) & 23 [11, 41] & 52.9\% [41.3, 64.8] \\
Verified 20{,}480 & 140/169 (82.8\%) & 10 [8, 19] & 45.2\% [33.1, 52.0] \\
Pro 49{,}152 & 246/316 (77.8\%) & 13 [8, 22] & 64.4\% [55.6, 71.3] \\
FeatureBench 47{,}104 & 143/183 (78.1\%) & 9 [7, 18] & 55.7\% [42.6, 64.6] \\
\bottomrule
\end{tabular*}
\end{table}

The tight-window runs shortened early and often, while the milder Verified
comparison shortened later and in fewer runs.
Appendix~Figure~\ref{fig:ctxcurves} follows six more task pairs.

\FloatBarrier
\subsection{What counts as success}
\label{sec:design:outcomes}

We report two complementary outcomes across the paired task sets. Resolution
records whether the benchmark evaluator accepts the whole task as solved,
while F2PF measures how much of the required behavior a patch repaired,
including on tasks that did not fully resolve.

Evaluators define \emph{fail-to-pass} (F2P) tests as tests that fail before a
patch and should pass after it. We call the share of this set that passes after
the patch the \emph{F2P fraction}, or F2PF:
\[
  \mathrm{F2PF}
  = \frac{\text{F2P tests made to pass}}
           {\text{all F2P tests}}.
\]

\begin{figure}[H]
\centering
\includegraphics[width=\linewidth]{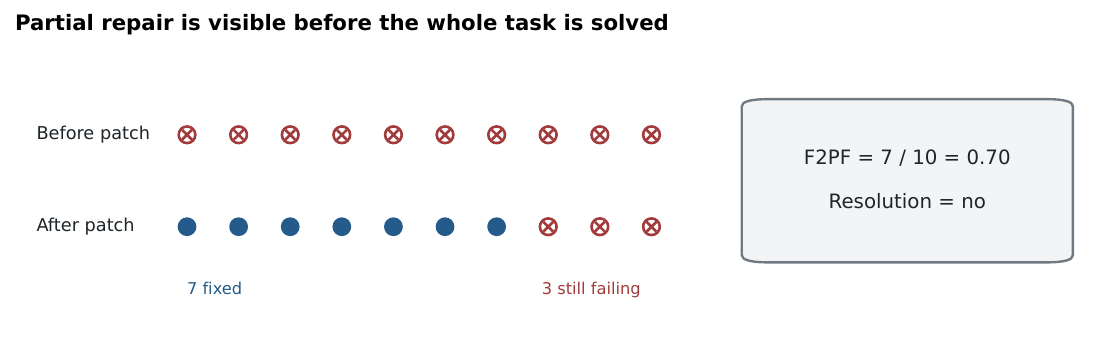}
\caption{The ten-test example. The patch fixes seven target tests, so its
F2PF is 0.70. Three target tests still fail, so the task is not resolved.
F2PF records useful movement below the all-or-nothing finish line.}
\label{fig:f2pf-example}
\end{figure}

Figure~\ref{fig:f2pf-example} shows how F2PF records partial repair without
turning it into complete resolution.

\FloatBarrier

For a finalized record with no F2P denominator, the fixed scoring rule sets
operational F2PF to zero in the task mean and paired difference. A measured
zero instead has a positive denominator and no passing target tests
(Figure~\ref{fig:f2pdist}). Each task contributes one operational F2PF score
between 0 and 1. We average these scores separately for control and treatment,
so every task has equal weight regardless of its number of tests. FeatureBench
can collect different target-test counts from two patches, so we use the
fraction that its evaluator reports for each task and arm. The task-level
outcomes for the pressure comparisons appear in the public
\href{https://github.com/sydches/yuj/blob/main/paper/results/task_outcomes.tsv}
{\texttt{task\_outcomes.tsv}} file, which lists every selected numerator and
denominator.

\subsection{How we compare the paired results}
\label{sec:design:stats}

We compare control and treatment on two task-level outcomes and use a different
paired test for each. Resolution is binary, so we use the exact two-sided
McNemar test to compare tasks resolved only by treatment with tasks resolved
only by control. F2PF is a quantity between zero and one, so we subtract control
from treatment for each task. A positive difference favors treatment, a
negative difference favors control, and zero is a tie. We set the ties aside
and use the exact two-sided sign test to ask which direction occurs more often.

The sign test gives each non-tied task one count, regardless of its F2P
denominator, but discards the size of its difference. Its p-value therefore
tests direction, while the reported arm means describe magnitude. A Wilcoxon
signed-rank test would rank the difference sizes, while a paired t-test would
test the mean difference. We chose the sign test because the frozen hypothesis
concerns direction across tasks, and the test requires neither a normal nor a
symmetric distribution of difference magnitudes. Each p-value applies to one
named frozen cohort, and we analyze benchmarks separately.

Because tasks from the same repository can share structure, we also report a
repository-level sensitivity analysis for the three primary pressure
comparisons. We first average the paired differences within each repository,
then apply an exact two-sided sign test to the non-tied repository directions.
We build the accompanying 95\% intervals from 20{,}000 resamples of whole
repositories with replacement, using seed 20260815. The task-paired tests
remain the primary analysis.

\subsection{Measuring context pressure}

We use several context-window sizes because the growing transcript creates the
pressure that motivates the treatment. After each comparison, we measure that
pressure as the share of initial control runs that fill their window. A high
share means that the control transcript regularly runs out of room, and using
the control arm gives each paired comparison one common measure.

We also ask how much reading a task requires before the model first changes
source code. An earlier large-window control run supplies this amount in prompt
tokens before the first write, and dividing it by the smaller pressured window
expresses reading demand as a share of the space available to the model.

For control and treatment separately, we then ask how likely a run is to fill
its window before making any source change. We call this event \emph{stopping
while reading}. The \emph{reading boundary} is the point where half the fitted
runs would stop this way. If the treatment boundary moves higher, the solver
can get through more reading before it runs out of room, while resolution and
F2PF still tell us whether it completed the work.
Appendix~\ref{app:experimental-custody} gives the fit and uncertainty
calculation.

\subsection{The same package across model designs}

After fixing the treatment settings in the Qwen3.6 study, we hold the package
fixed while changing the model. Each model has a basic compatibility setup
shared by its control and treatment runs, but we do not retune the treatment
rules. The registered comparisons use the same 169 Verified tasks,
20{,}480-token window, fixed 480-second attempt budget, and frozen treatment
package. We report each model separately under its registered analysis family,
with exact model identities and designs in Section~\ref{sec:validation}.

% Primary Qwen3.6 results. Detailed mechanism and execution tables are in
% Appendix A.

\section{Changing the harness changed the results}
\label{sec:dose}

\subsection{Progress and complete solutions under context pressure}

The largest changes appear when the growing control transcript presses against
the context limit. Mean per-task F2PF increases under treatment on all three
pressured cohorts, and each exact paired sign test gives $p<0.0001$.
Complete solutions move in the same direction on the two pressured SWE-bench
cohorts, and both exact paired McNemar tests give $p<0.0001$. On the
\QFullFBPressN{}-task FeatureBench pressure cohort, mean per-task F2PF rises
from \QFullFBPressFtpC\% to \QFullFBPressFtpT\%, while complete solutions
remain sparse at \QFullFBPressResC{} under control and \QFullFBPressResT{}
under treatment. After Holm correction across the six repository-level tests,
all three F2PF increases and the Pro resolution increase remain significant at
0.05, while the adjusted resolution tests for Verified and FeatureBench are
not significant
(Appendix~Table~\ref{tab:qwen-cluster-sensitivity}).

Table~\ref{tab:qwen-pressure-summary} summarizes the outcomes for the
three primary pressure comparisons. The complete table in the appendix adds
the larger context regimes and paired tests.

% AUTO-GENERATED by utils/make_qwen_outcomes.py
\begin{table}[H]
\centering
\caption{Outcomes for the three primary Qwen3.6 pressure comparisons. Each row gives mean per-task F2PF and complete solutions for both arms. We round percentages to whole numbers. C and T denote control and treatment.}
\label{tab:qwen-pressure-summary}
\small
\begin{tabular*}{\linewidth}{@{\extracolsep{\fill}}lcc@{}}
\toprule
benchmark & \shortstack{mean per-task F2PF\\C $\rightarrow$ T} & \shortstack{complete solutions\\C $\rightarrow$ T} \\
\midrule
SWE-bench Verified & 28\% $\rightarrow$ 49\% & 43 $\rightarrow$ 72 \\
SWE-bench Pro & 15\% $\rightarrow$ 33\% & 31 $\rightarrow$ 72 \\
FeatureBench & 11\% $\rightarrow$ 20\% & 2 $\rightarrow$ 3 \\
\bottomrule
\end{tabular*}
\end{table}

\FloatBarrier

Less model work can mark an earlier stop rather than greater efficiency: under
pressure, control often reached the context limit while treatment continued.
Appendix~Table~\ref{tab:pressure-resources} reports total model turns, prompt
tokens, and solver wall time for each pressure comparison.

\subsection{Results across context regimes}

\begin{figure}[H]
\centering
\includegraphics[width=\linewidth]{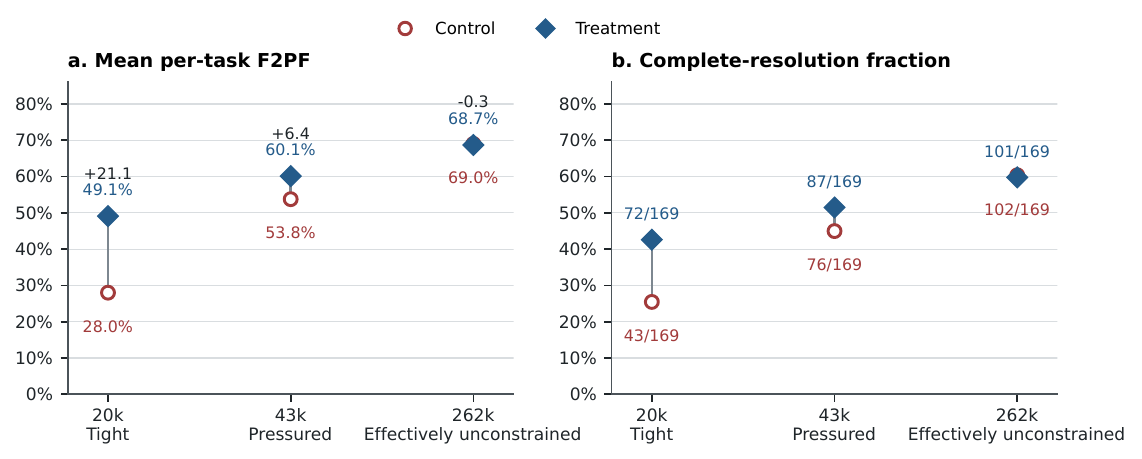}
\caption{Qwen3.6 outcomes across the three Verified context windows at the
completed-run endpoint under the fixed 480-second budget. Every point uses the
same \QGradN{} tasks. The mean per-task F2PF label above each window gives the
paired treatment-minus-control difference in percentage points.}
\label{fig:window-gradient}
\end{figure}

Figure~\ref{fig:window-gradient} plots outcomes at three Verified windows. All
three comparisons use the same \QGradN{} task IDs and fixed 480-second budget;
this common cohort is the set of tasks from the original 187-task pool that
produced paired outcomes within the budget. Each point shows one arm's outcome
at the completed-run endpoint. Across these observed windows, control scores
are higher at larger windows, while the treatment-control gap is smaller for
both mean per-task F2PF and complete solutions. The gap is largest at
20{,}480 tokens and closest to zero at 262{,}144 tokens. Because we collected
the campaigns successively, the cross-window ordering does not isolate window
size from run-era change.
Appendix~Table~\ref{tab:verified-intervals} gives the paired intervals.

At 43{,}008 tokens, the comparison remains positive on both outcomes: the
bootstrap intervals for the mean differences exclude zero, while the exact
paired direction tests lie just above 0.05. Appendix~Table~\ref{tab:qwen-full-outcomes}
reports the outcomes and exact tests, and Appendix~Table~\ref{tab:verified-intervals}
reports the intervals.

At 262{,}144 tokens, context was effectively unconstrained across all three
cohorts: in every benchmark and arm, fewer than 1\% of tasks hit the allotted
context. On Verified and Pro, the treatment advantage seen under pressure
largely disappears: mean per-task F2PF and complete solutions are close between
arms. On Verified, the treatment-minus-control mean per-task F2PF difference
was \QGradWideFtpDiff{} percentage points (95\% interval
[\QGradWideFtpDiffCILo{}, \QGradWideFtpDiffCIHi{}]). Treatment also served
fewer prompt tokens per turn, and neither arm ended at the context limit. On
Verified, the wide-window result therefore shows that treatment can reduce the
working view while observed outcomes remain close.
Appendix~Tables~\ref{tab:verified-intervals}
and~\ref{tab:verified-wide-efficiency} report the paired intervals and context
use.

FeatureBench differs. On the same \QFullFBWideN{} tasks, mean per-task F2PF
rises from \QFullFBWideFtpC\% to \QFullFBWideFtpT\%. The primary task-paired
sign test gives $p=\QFullFBWideFtpP{}$, while the repository-level direction
sign test gives $p=0.0963$. Complete solutions remain \QFullFBWideResC{} in
control and \QFullFBWideResT{} in treatment.

\subsection{F2PF movement across both regimes}

\begin{figure}[t]
\centering
\includegraphics[width=\linewidth]{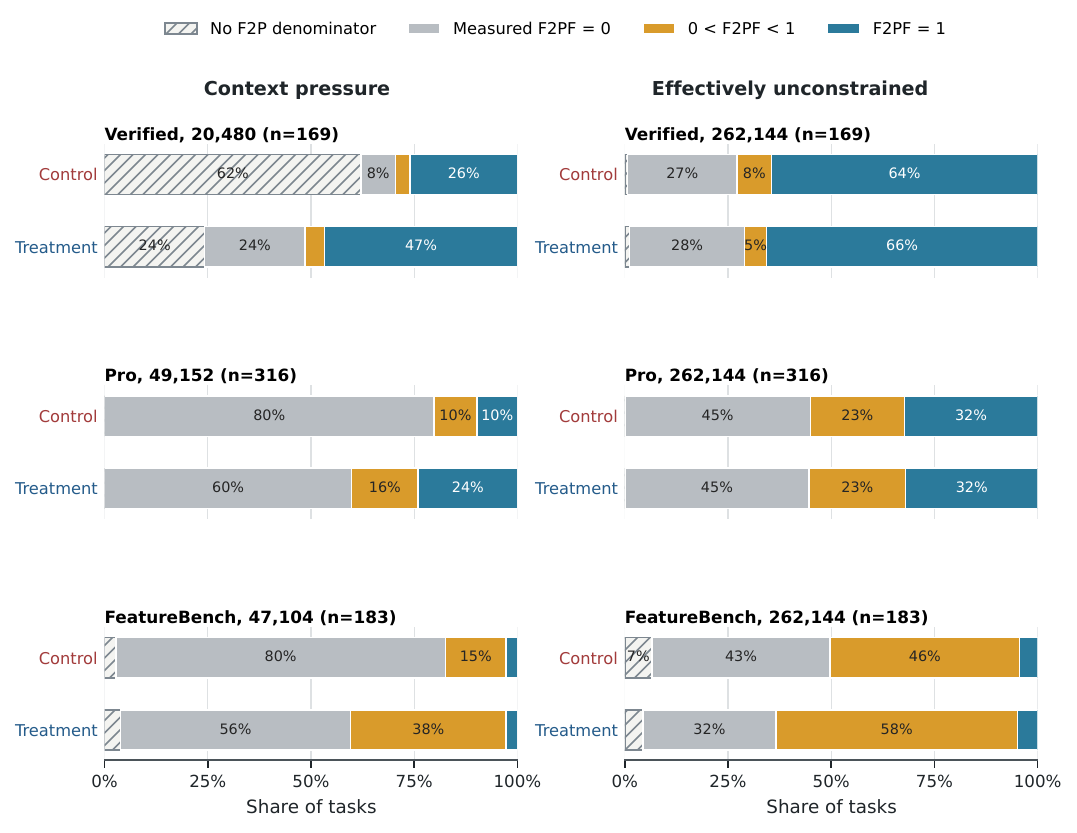}
\caption{One operational F2PF state per task under context pressure and with
context effectively unconstrained at 262{,}144 tokens. Hatched
segments show selected records with no F2P denominator. The fixed scoring rule
assigns those records zero. Gray segments show measured zeros with a positive
denominator. We round segment percentages independently. Each panel uses the
task outcomes for the matching benchmark and window in
Appendix~Table~\ref{tab:qwen-full-f2pf}.}
\label{fig:f2pdist}
\end{figure}

F2PF shows what happened below the all-or-nothing finish line. Each task,
resolved or unresolved, contributes one F2PF value per arm, so the state
distribution describes movement across the full cohort.
Figure~\ref{fig:f2pdist} shows the F2PF state distribution in both context
regimes.

Under tight context limits, the share of tasks assigned an operational zero
is lower with treatment in all three pressure comparisons. Verified moves
mainly toward complete test repair, FeatureBench moves mainly into the middle
where some target tests pass but the whole task is not yet solved, and Pro
shows both kinds of movement. At 262{,}144 tokens, the Verified and Pro
distributions remain close, while FeatureBench keeps a visible shift toward
partial and complete test repair.
\FloatBarrier

\subsection{The treatment moved the reading boundary}

\begin{figure}[H]
\centering
\includegraphics[width=\linewidth]{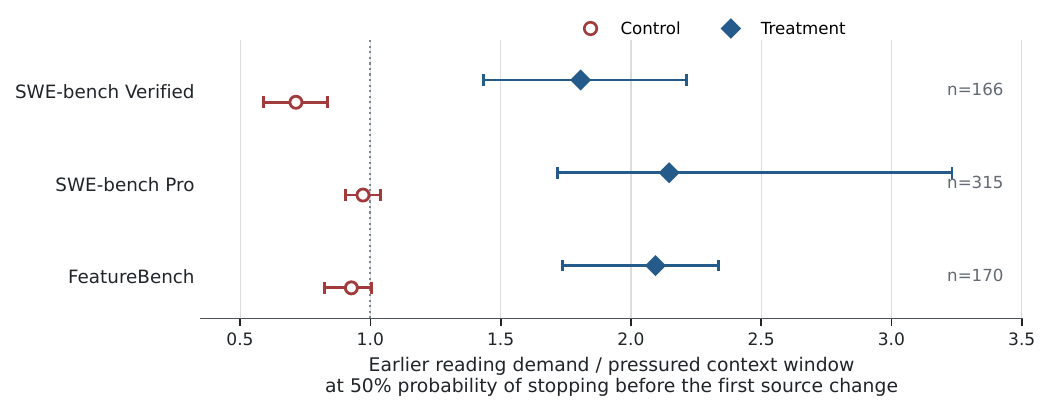}
\caption{The fitted reading boundary for the three Qwen3.6 pressure
comparisons. Horizontal lines are 95\% repository-cluster bootstrap
intervals, and the dotted line marks demand equal to one pressured window.
A point farther right means that the model can complete more reading before
half the runs stop.}
\label{fig:reading-boundary}
\end{figure}

The paired outcomes show when the treatment matters, while the reading analysis
examines one mechanism active under pressure: some tasks require substantial
reading before the first source change. Figure~\ref{fig:reading-boundary} shows
the fitted point where half the runs would stop during that reading. On every
benchmark, treatment moved this boundary to more than twice the control
location, and every repository-bootstrap interval for that movement excludes
one.

\FloatBarrier

Moving the reading boundary does not by itself resolve a task. Resolution and
F2PF remain the outcome measures.

\section{The same treatment across model designs}
\label{sec:validation}

We held the frozen treatment package fixed while changing the model design.
The paired comparisons use the same Verified cohort, tight context window, and
final within-budget endpoint as the Qwen3.6 pressure study, while both arms
within each model use the same declared four-bit weight file and compatibility
setup. The transfer models required model-specific profiles and serving
settings, but no harness-code fork or treatment retuning.

Table~\ref{tab:crossmodel-outcomes} places these paired measurements beside
the primary-model pressure result. Treatment raises mean per-task F2PF and
complete solutions for every tested model, with Devstral showing the largest
proportional changes: 2.1$\times$ the control mean per-task F2PF and
2.4$\times$ as many complete solutions on the 169-task Verified cohort at the
20{,}480-token window and fixed 480-second attempt endpoint.
Figure~\ref{fig:window-gradient} follows Qwen3.6 across context windows,
whereas Table~\ref{tab:crossmodel-outcomes} holds the window fixed and compares
model designs.

% AUTO-GENERATED by pipelines/make_result_tables.py
\begin{table}[H]
\centering
\small
\caption{Paired outcomes on the same 169-task SWE-bench Verified cohort at 20{,}480 tokens and a fixed 480-second attempt budget. C and T denote control and treatment. The design column uses the term mixture of experts (MoE). Parentheses give the absolute gain and T/C multiplier; F2PF gains are percentage points.}
\label{tab:crossmodel-outcomes}
\begin{tabular*}{\linewidth}{@{\extracolsep{\fill}}llll@{}}
\toprule
model & design & F2PF C $\rightarrow$ T (gain; $\times$) & solutions C $\rightarrow$ T (gain; $\times$) \\
\midrule
Qwen3.6~\cite{qwen36_35b_a3b} & DeltaNet/attention MoE & 28\% $\rightarrow$ 49\% (+21; 1.8$\times$) & 43 $\rightarrow$ 72 (+29; 1.7$\times$) \\
Devstral~\cite{devstral_small2} & dense transformer & 17\% $\rightarrow$ 37\% (+20; 2.1$\times$) & 22 $\rightarrow$ 53 (+31; 2.4$\times$) \\
Nemotron~\cite{nemotron_cascade2} & Mamba-2/attention MoE & 12\% $\rightarrow$ 18\% (+6; 1.5$\times$) & 16 $\rightarrow$ 25 (+9; 1.6$\times$) \\
Qwen3.8~\cite{qwen38_27b} & dense DeltaNet/attention & 20\% $\rightarrow$ 35\% (+15; 1.7$\times$) & 32 $\rightarrow$ 54 (+22; 1.7$\times$) \\
\bottomrule
\end{tabular*}
\end{table}

The exact paired tests and applicable correction appear in
Appendix~Table~\ref{tab:crossmodel-full187}.

\FloatBarrier
\section{What these results change}
\label{sec:discussion}

\subsection{The tested solver is the model and harness together}

A model name alone does not identify the solver behind measured coding-agent
performance. The model proposes every search, edit, and test, while the harness
shapes the conditions in which it does so. In the paired results, the model
weights and tasks stayed fixed while changing the harness changed the completed
work, so the result belongs to the model and harness together. Two evaluations
that name the same model can therefore measure different solvers if their
harnesses or harness configurations differ.

\subsection{Harness design is another lever}

The observed arm gap differs across context regimes: treatment produces large
gains on all three pressure cohorts, whereas observed paired outcomes are close
on Verified and Pro when context is effectively unconstrained. FeatureBench
retains a mean per-task F2PF gain in that regime; the gain is significant in
the primary task-paired test, but not in the repository-level direction sign
test.

Long coding tasks accumulate full files, search results, command output,
and repeated tests, so that history competes with the new evidence needed for
the next decision. A harness can keep the full record while treating the
model's current context as a mechanically constructed working set rather than
an ever-growing archive. The working-view rule starts shortening old results
only when context pressure develops.

We cannot know at the start of a long task whether context will bind. Treatment
produced large gains when it did, while observed performance remained close on
Verified and Pro when context remained ample. This pattern supports enabling
the treatment from the start, whether or not context later binds.

\subsection{Implications for builders and evaluators}

Builders should treat view construction, tool behavior, and run control as
parts of solver design, and should version and test these choices as carefully
as the model setup. Evaluations should report the model, harness, tool
interface, model-visible context, context limit and policy, time limits, and
the rule used to choose the final result.

\subsection{Scope of the evidence}

The comparison estimates the treatment-package effect as run, so differences
in realized model work after control reached its context limit are part of that
effect rather than a compute-matched efficiency estimate. Each task contributes
one greedy trajectory per arm, so this design does not measure run-to-run
variation.

The outcome evidence covers four locally served open-weight checkpoints, each
using its declared four-bit weight file; higher-precision weights and hosted
frontier models lie outside the tested set. All three benchmarks use
repository-level coding tasks, and success means the benchmark evaluator's
test-based verdict and per-task F2PF. Human patch-quality judgment was not an
endpoint, and the study does not test other task types.

Both paired arms use the same model weights and therefore share any benchmark
exposure in pretraining. This pairing prevents arm assignment from changing
benchmark exposure, but the design does not measure interactions between that
exposure and the treatment. The chosen control sets another boundary: the
experiment compares the mechanical treatment with the full
chronological-transcript control, not with model-written summaries or
retrieval-based views.

\section{What to test next}
\label{sec:future}

The present study fixes the treatment package and time budget within each
comparison, and each task contributes one trajectory per arm. The next
experiments should first separate those limits, then test how the complete
harness adapts across tasks and model designs.

\subsection{What does the detector response add?}

The paired results estimate the effect of the complete treatment package, not
any one rule. A prespecified component study should hold half-life context and
command safeguards fixed while varying only whether the harness delivers a
detector response. This comparison would measure what the detector response
adds to the fixed package.

\subsection{What changes when only time changes?}

One experiment should hold the model, tasks, tools, and context policy fixed
while varying the available time. Repeated runs at several limits would
separate tasks that need more time from those that need more context.

\subsection{Can the complete harness adapt to the task?}

One experiment should let the complete harness adjust to the state
of a run. For example, the harness could respond to how quickly the context is
filling and whether the model has begun to edit. We should fix the rule before
testing and then use it on new tasks. Repeated control and treatment runs would
show the average change and the run-to-run variation.

A simple extension is to cache repeated read-only results. Before reusing
a saved result, the harness would confirm that the request and the source hash
match their saved values. If neither has changed, it would return a short
reference with the
beginning and end of the saved result instead of adding the full output
again. If the source has changed, or the model asks for a fresh read, the
harness would run the request normally. This would prevent duplicate output
mechanically, before the context policy has to shorten it.

\subsection{How does model design interact with context?}

The cross-model results motivate a direct study of how model design interacts
with context pressure. We should compare the same cohort at severe, moderate,
and unconstrained windows. This comparison would show how quickly each model
approaches its own ceiling and how much of the pressure gap the harness
recovers. A direct study could start with Devstral because treatment produces
greater-than-twofold increases in both mean per-task F2PF and complete
solutions on the 169-task Verified cohort at the 20{,}480-token window and
fixed 480-second attempt endpoint.

\section{Conclusion}
\label{sec:conclusion}

The same model produced different coding-agent results under different
configurations of the same harness. Under context pressure, treatment repaired
more required behavior on all three benchmarks and completed more tasks on
Verified and Pro. When context was effectively unconstrained, observed outcomes
were close on Verified and Pro, while FeatureBench retained a mean per-task
F2PF gain that was significant in the primary task-paired test but not in the
repository-level direction sign test. On the same 169-task Verified cohort at
the 20{,}480-token window and fixed 480-second attempt endpoint, treatment
raised mean per-task F2PF and complete solutions for every tested model,
although the gain differed by model.

Long repository tasks collect files, searches, commands, and test output, but a
harness can preserve that record while managing the model's working view and
responding to the run as it develops. The paired results therefore support
treating the model and harness as one solver. Builders should design and test
that complete system, and evaluations should report the model, harness, tool
interface, context policy, and run controls.

\clearpage
\begin{appendices}
\section{Methods and supporting analyses}
\label{app:experimental-custody}

\subsection{Outcome selection}

Each task contributes one finalized result per arm under the comparison's fixed
attempt budget and scoring protocol. The endpoint is the best finalized task
outcome produced under that protocol.

\subsection{Cohorts and evaluation setup}

Table~\ref{tab:cohort-rules} states the benchmark-specific cohort rule for each
paired comparison.

\begin{table}[H]
\centering
\small
\caption{Cohorts used in the paired comparisons.}
\label{tab:cohort-rules}
\begin{tabular*}{\linewidth}{@{\extracolsep{\fill}}lrp{0.66\linewidth}@{}}
\toprule
benchmark & $n$ & selection rule \\
\midrule
SWE-bench Verified & \VCohortN{} & The original pressure pool contained 187
tasks. We applied the fixed 480-second cap to both arms and retained the 169
tasks that produced complete paired outcomes within that cap. We use the same
task IDs at every Verified window and on every model. \\
SWE-bench Pro & \ProCN{} & We select tasks whose earlier large-window control
run crossed the frozen reading-demand threshold. \\
FeatureBench & \FBN{} & We use every canonical task runnable in the evaluation
setup. Reading demand does not determine membership. \\
\bottomrule
\end{tabular*}
\end{table}

The primary Qwen3.6 runs use \texttt{llama-server}, greedy decoding, and
four-bit Q4\_K\_XL weights. Fixed analysis rules select one finalized evaluator
record per task and assign outcomes for evaluator failures, empty collected
patches, and wrong-window records.

Each task contributed one independently executed trajectory per arm under
greedy decoding, so the study estimates the task-paired difference under the
fixed execution protocol rather than pass@$k$ or a task-specific stochastic
success probability. The exact paired tests therefore treat tasks, not
resampled trajectories, as the units of inference. This design does not
measure run-to-run variation or end-to-end determinism, which require a
repeated-run design.

Table~\ref{tab:verified-matched-window-design} records the conditions shared
across the three Verified window comparisons.

\begin{table}[H]
\centering
\small
\caption{Fixed design for the matched-window Verified comparison.}
\label{tab:verified-matched-window-design}
\begin{tabular*}{\linewidth}{@{\extracolsep{\fill}}lp{0.72\linewidth}@{}}
\toprule
item & fixed setting \\
\midrule
Cohort & We use the same \VCohortN{} task IDs at every window. \\
Endpoint & We select the best finalized patch within the fixed 480-second
budget. \\
Context windows & We use 20{,}480, 43{,}008, and 262{,}144 tokens. \\
Within a campaign & Control and treatment share one model-facing code state. \\
Across windows & We collected campaigns successively. The ordering describes
the observed context regimes. \\
\bottomrule
\end{tabular*}
\end{table}

Table~\ref{tab:verified-intervals} reports the paired mean-difference intervals
for those three comparisons.

% AUTO-GENERATED by pipelines/make_result_tables.py
\begin{table}[H]
\centering
\small
\caption{Completed-run paired differences on the same 169-task Verified cohort under the fixed 480-second budget. Each entry gives treatment minus control in percentage points, with paired-task bootstrap 95\% intervals.}
\label{tab:verified-intervals}
\begin{tabular*}{\linewidth}{@{\extracolsep{\fill}}lrr@{}}
\toprule
window & F2PF difference {[}95\% interval{]} & resolution difference {[}95\% interval{]} \\
\midrule
20{,}480 (tight) & +21.1 [+14.1, +28.3] & +17.2 [+10.7, +23.7] \\
43{,}008 (pressured) & +6.4 [+0.5, +12.4] & +6.5 [+0.6, +13.0] \\
262{,}144 (effectively unconstrained) & -0.3 [-4.5, +3.9] & -0.6 [-5.9, +4.7] \\
\bottomrule
\end{tabular*}
\end{table}

\FloatBarrier

Tables~\ref{tab:qwen-full-outcomes} and~\ref{tab:qwen-full-f2pf} add the exact
paired outcome tests and the other observed context regimes.

% AUTO-GENERATED by utils/make_qwen_outcomes.py
\begin{table}[H]
\centering
\caption{Complete solutions in the paired Qwen3.6 comparisons. C and T denote control and treatment. The pressure rows are primary. The remaining rows report observed results at larger context windows. Tests compare arms within each row. The Verified wide-window row is a post-hoc common-cohort projection.}
\label{tab:qwen-full-outcomes}
\small
\setlength{\tabcolsep}{3pt}
\begin{tabular}{llrrrl}
\toprule
benchmark & window & $n$ & solutions C & solutions T & T-only:C-only ($p$) \\
\midrule
SWE-bench Verified & 20{,}480 & 169 & 43 & 72 & 34:5 ($<0.0001$) \\
SWE-bench Pro & 49{,}152 & 316 & 31 & 72 & 47:6 ($<0.0001$) \\
FeatureBench & 47{,}104 & 183 & 2 & 3 & 1:0 ($1$) \\
\addlinespace
SWE-bench Verified & 43{,}008 & 169 & 76 & 87 & 20:9 ($0.0614$) \\
\addlinespace
SWE-bench Verified (common cohort) & 262{,}144 & 169 & 102 & 101 & 9:10 ($1$) \\
SWE-bench Pro & 262{,}144 & 316 & 101 & 99 & 11:13 ($0.8388$) \\
FeatureBench & 262{,}144 & 183 & 5 & 5 & 1:1 ($1$) \\
\bottomrule
\end{tabular}
\end{table}

\begin{table}[H]
\centering
\caption{Mean per-task F2PF in the same paired Qwen3.6 comparisons. C and T denote control and treatment. The final column gives task-level direction and its exact paired sign-test $p$-value.}
\label{tab:qwen-full-f2pf}
\small
\setlength{\tabcolsep}{3pt}
\begin{tabular}{llrrrl}
\toprule
benchmark & window & F2PF C (\%) & F2PF T (\%) & ratio & higher T:C:tie ($p$) \\
\midrule
SWE-bench Verified & 20{,}480 & 28.0 & 49.1 & 1.75 & 42:6:121 ($<0.0001$) \\
SWE-bench Pro & 49{,}152 & 15.2 & 32.7 & 2.15 & 84:13:219 ($<0.0001$) \\
FeatureBench & 47{,}104 & 10.5 & 19.6 & 1.86 & 47:9:127 ($<0.0001$) \\
\addlinespace
SWE-bench Verified & 43{,}008 & 53.8 & 60.1 & 1.12 & 21:10:138 ($0.0708$) \\
\addlinespace
SWE-bench Verified (common cohort) & 262{,}144 & 69.0 & 68.7 & 1.00 & 9:8:152 ($1$) \\
SWE-bench Pro & 262{,}144 & 44.8 & 44.5 & 0.99 & 22:24:270 ($0.883$) \\
FeatureBench & 262{,}144 & 23.9 & 30.7 & 1.28 & 61:26:96 ($0.00022$) \\
\bottomrule
\end{tabular}
\end{table}

\FloatBarrier

The three Verified rows use the current frozen treatment package. The Pro and
FeatureBench pressure rows use the earlier complete version frozen for those
campaigns. Their 262{,}144-token rows are historical comparisons rather than
exact final-package replications. The FeatureBench wide-window row pairs an
earlier treatment with a later corrected control.

The public
\href{https://github.com/sydches/yuj/blob/main/paper/provenance/cell_provenance.json}
{\texttt{cell\_provenance.json}} record identifies all seven Qwen3.6 rows. It
gives resolved-configuration hashes for all four primary pressure cells and
complete code identity for three. For Pro, 632 of 871 finalized opportunities
have independent code stamps; 239 later continuation opportunities do not. The
missing stamps limit the code-identity record but do not show that different
code ran.

FeatureBench reports each evaluator's task-level fraction even when the two
patches collect different target-test counts. The public
\href{https://github.com/sydches/yuj/blob/main/paper/results/task_outcomes.tsv}
{\texttt{task\_outcomes.tsv}} file gives both values for every selected record.
As a denominator-size sensitivity check, we divided each arm's passed count by
the larger observed arm denominator for that task. In the FeatureBench pressure
comparison, the resulting means were \QFullFBPressCommonDenomFtpC\% for
control and \QFullFBPressCommonDenomFtpT\% for treatment. The non-tied
direction count remained
\QFullFBPressCommonDenomFtpTHigher:\QFullFBPressCommonDenomFtpCHigher{} in
favor of treatment. Unequal denominator size therefore does not explain the
observed direction.

\subsection{Resource use under pressure}

Table~\ref{tab:pressure-resources} reports the model work consumed by the same
final endpoints as the three primary Qwen3.6 pressure comparisons. Treatment
used more model work in every comparison and more often remained active after
control reached its context limit. The paired efficacy results include this
difference in realized work.

% AUTO-GENERATED by pipelines/make_resource_accounting.py
\begin{table}[H]
\centering
\caption{Recorded model work for the three primary Qwen3.6 pressure comparisons. Each total covers the complete fixed run protocol used for the final outcome. Model turns count unique recorded turns within each solver invocation, while prompt tokens count the assembled model inputs at those turns. C/T denotes control/treatment.}
\label{tab:pressure-resources}
\small
\setlength{\tabcolsep}{3pt}
\begin{tabular*}{\linewidth}{@{\extracolsep{\fill}}lrrrr@{}}
\toprule
benchmark & $n$ & \shortstack[r]{model turns\\C / T} & \shortstack[r]{prompt tokens (millions)\\C / T} & \shortstack[r]{solver wall time (hours)\\C / T} \\
\midrule
SWE-bench Verified & 169 & 3{,}280 / 6{,}517 & 37.8 / 80.7 & 1.3 / 4.6 \\
SWE-bench Pro & 316 & 8{,}357 / 25{,}750 & 210.6 / 657.1 & 5.4 / 21.8 \\
FeatureBench & 183 & 5{,}281 / 15{,}178 & 140.3 / 412.8 & 2.5 / 14.7 \\
\bottomrule
\end{tabular*}
\end{table}

\FloatBarrier

\subsection{Wide-window context use}

Table~\ref{tab:verified-wide-efficiency} separates two parts of context use in
the 262{,}144-token Verified comparison. Prompt tokens per turn measure the
working-view compression applied by the treatment, whereas prompt tokens per
task also reflect how many turns the run took. Among the
\QFullVWideCompressionN{} paired tasks with defined prompt-token-per-turn
ratios, the geometric mean shows that treatment served
\QFullVWideCompressionReduction\% fewer prompt tokens per turn; among the
\QFullVWideServedN{} paired tasks with both arm metrics, treatment's mean number
of turns was \QFullVWideTurnsIncrease\% higher.

% AUTO-GENERATED by utils/make_qwen_outcomes.py
\begin{table}[H]
\centering
\caption{Context use in the 262{,}144-token Verified comparison. Each point estimate is the geometric mean of paired treatment/control task ratios. We build the intervals from 10{,}000 paired task-bootstrap samples. A ratio below one means that treatment served less context.}
\label{tab:verified-wide-efficiency}
\small
\setlength{\tabcolsep}{4pt}
\begin{tabular*}{\linewidth}{@{\extracolsep{\fill}}lrr@{}}
\toprule
measure & paired tasks & T/C ratio {[}95\% interval{]} \\
\midrule
Prompt tokens per task & 154 & 0.959 [0.856, 1.071] \\
Prompt tokens per turn & 153 & 0.928 [0.878, 0.979] \\
\bottomrule
\end{tabular*}
\end{table}

\FloatBarrier

\subsection{Cross-model paired statistics}

Table~\ref{tab:crossmodel-full187} reports the paired estimates, intervals, and
exact tests for the fixed-window transfer comparisons.

% AUTO-GENERATED by pipelines/make_result_tables.py
\begin{table}[H]
\centering
\small
\caption{Transfer-model outcome comparisons on the frozen 169-task SWE-bench Verified pressure cohort at a 20{,}480-token window and the fixed 480-second attempt endpoint. F2PF differences are treatment minus control in percentage points. Paired task bootstraps give 95\% intervals in the same units. Exact paired sign tests compare F2PF directions, while exact paired McNemar tests compare resolution as a secondary outcome. The Holm column applies to the pre-registered Devstral--Nemotron family, whereas the separately registered Qwen3.8 comparison is outside that family.}
\label{tab:crossmodel-full187}
\begin{tabular}{lrrrrrr}
\toprule
model & F2PF C/T (\%) & \shortstack[r]{difference\\{[}95\% interval{]}} & $p$ & $p_{\mathrm{Holm}}$ & solutions C/T & $p_{\mathrm{McNemar}}$ \\
\midrule
Devstral & 17.2/36.8 & +19.5 [+12.0, +26.9] & $<0.0001$ & $<0.0001$ & 22/53 & $<0.0001$ \\
Nemotron & 12.0/18.3 & +6.4 [+1.4, +11.4] & $0.0169$ & $0.0169$ & 16/25 & $0.0636$ \\
Qwen3.8 & 20.4/35.3 & +14.9 [+9.9, +20.4] & $<0.0001$ & -- & 32/54 & $<0.0001$ \\
\bottomrule
\end{tabular}
\end{table}

\FloatBarrier

The reported transfer comparisons use the same frozen 169-task manifest,
20{,}480-token window, 480-second attempt budget, and treatment package.
Within each reported model, control and treatment use the same declared
four-bit weight file, serving settings, chat template, solver tree, and task
images. All sessions selected for the reported rows pass the registered
serving-contract and content-equivalence checks.

\subsection{Task isolation}
\label{app:task-isolation}

For every paired comparison, control and treatment began from separate copies
of the same prepared task checkout and used the same isolation settings. The
solver ran commands inside per-task containers with outbound network access
disabled. The solver therefore could not fetch from GitHub, package indexes, or
other remote services. The containers did not mount benchmark task tables,
gold patches, or scorer-only inputs during model steps.

Verified used locally sealed task images. Preparation replaced the repository
history with one synthetic base commit and removed remotes, tags, reflogs, and
unreachable Git objects. It also masked other installed copies of the target
package, caches, credentials, and stray patch or diff files. A fail-closed
preflight checked the base commit and the non-root container. It also checked
that the container had no Docker socket, DNS, or direct-socket egress before
the model began work.

FeatureBench applied the benchmark mask, removed the hidden F2P tests and their
bytecode, and rebuilt the checkout as one parentless task commit. The hidden
test patch remained outside the model-visible checkout. The harness mounted it
only for scoring steps. The container root was read-only. Only the prepared task
tree was writable.

For SWE-bench Pro, we copied both arms byte-for-byte from the same benchmark
task image before they ran. The benchmark task table and gold-patch files
remained outside the container mounts. The Pro task image retained its upstream
Git metadata.

\subsection{Repository-cluster sensitivity}

Tables~\ref{tab:qwen-cluster-sensitivity}
and~\ref{tab:qwen-cluster-resolution-sensitivity} repeat the three primary
pressure comparisons with repositories as the unit of direction and
resampling.

% AUTO-GENERATED by utils/make_qwen_outcomes.py
\begin{table}[H]
\centering
\caption{Repository-cluster sensitivity of mean per-task F2PF in the three primary Qwen3.6 pressure comparisons. Differences are treatment minus control in percentage points. The final column gives repository-level direction and the Holm-adjusted exact sign-test $p$-value. The adjustment covers the six F2PF and resolution tests.}
\label{tab:qwen-cluster-sensitivity}
\small
\setlength{\tabcolsep}{4pt}
\begin{tabular}{lrrl}
\toprule
benchmark & repositories & \shortstack{F2PF difference\\{[}95\% interval{]}} & higher T:C:tie ($p_{\mathrm{Holm}}$) \\
\midrule
SWE-bench Verified & 11 & 21.1 [13.7, 25.0] & 7:0:4 ($0.0469$) \\
SWE-bench Pro & 11 & 17.5 [11.8, 24.3] & 11:0:0 ($0.00586$) \\
FeatureBench & 22 & 9.1 [5.8, 14.1] & 16:2:4 ($0.00586$) \\
\bottomrule
\end{tabular}
\end{table}

\begin{table}[H]
\centering
\caption{Repository-cluster sensitivity of complete resolution in the same comparisons. Differences are treatment minus control in percentage points. The final column uses the same six-test Holm family as Table~\ref{tab:qwen-cluster-sensitivity}.}
\label{tab:qwen-cluster-resolution-sensitivity}
\small
\setlength{\tabcolsep}{4pt}
\begin{tabular}{lrrl}
\toprule
benchmark & repositories & \shortstack{resolution difference\\{[}95\% interval{]}} & higher T:C:tie ($p_{\mathrm{Holm}}$) \\
\midrule
SWE-bench Verified & 11 & 17.2 [9.6, 20.1] & 6:0:5 ($0.0625$) \\
SWE-bench Pro & 11 & 13.0 [8.2, 19.1] & 11:0:0 ($0.00586$) \\
FeatureBench & 22 & 0.5 [0.0, 1.5] & 1:0:21 ($1$) \\
\bottomrule
\end{tabular}
\end{table}

\FloatBarrier

\subsection{Detailed Qwen3.6 reading and treatment activity}

For Table~\ref{tab:shortening-activity}, an active run has at least one record
where shortening removed characters, and its typical reduction is the median
across exact-deduplicated events. The table then reports the median and middle
half across active runs.

Table~\ref{tab:reading-fit-specification} records the fit and uncertainty
calculation, and Table~\ref{tab:reading-pressure} gives the resulting reading
analysis behind Section~\ref{sec:dose}. For each benchmark, the fitted
treatment location is above the control location and the interval for its
treatment-to-control ratio excludes one. Because the control locations differ
across benchmarks, each printed location describes only its benchmark; these
window-scale results are descriptive.

\begin{table}[H]
\centering
\small
\caption{Reading-boundary fit and uncertainty calculation.}
\label{tab:reading-fit-specification}
\begin{tabular*}{\linewidth}{@{\extracolsep{\fill}}lp{0.72\linewidth}@{}}
\toprule
item & specification \\
\midrule
Fit & We fit a logistic regression of stopping while reading on
$\log_2(\text{demand}/\text{window})$. \\
Interval & We use the 2.5th to 97.5th percentile of the bootstrap distribution. \\
Resampling & We draw 2{,}000 repository samples with replacement, using seed
20260806. \\
Excluded draws & We exclude draws with one outcome class, a singular fit, or a
non-positive fitted slope. \\
Paired ratio & We use the same repository sample for both arms. \\
Valid ratios & All 2{,}000 paired samples produced estimates on all three
benchmarks. \\
\bottomrule
\end{tabular*}
\end{table}

\FloatBarrier
% AUTO-GENERATED by pipelines/make_result_tables.py
\begin{table}[H]
\centering
\small
\caption{Fitted Qwen3.6 reading boundaries. Each location gives the demand-to-window ratio where half the runs stop before the first source change. Brackets give repository-cluster bootstrap 95\% intervals. Stretch equals the treatment location divided by the control location.}
\label{tab:reading-pressure}
\begin{tabular}{lrrrr}
\toprule
benchmark & $n$ & control location & treatment location & stretch \\
\midrule
SWE-bench Verified & 166 & 0.71 [0.59, 0.84] & 1.81 [1.43, 2.21] & 2.53 [1.81, 3.59] \\
SWE-bench Pro & 315 & 0.97 [0.90, 1.04] & 2.15 [1.72, 3.23] & 2.21 [1.80, 3.25] \\
FeatureBench & 170 & 0.93 [0.82, 1.01] & 2.09 [1.74, 2.34] & 2.26 [1.86, 2.56] \\
\bottomrule
\end{tabular}
\end{table}

\FloatBarrier

Tables~\ref{tab:channel-context-activity}
and~\ref{tab:channel-detector-activity} report treatment events in the paired
populations: changes to the current context view, command safeguards, valid
detector decisions, and responses applied by the harness. An event appears
when its condition arises. Matching control sessions contain none of these
markers.

In the first table, \emph{command} counts rewritten command calls. The
\emph{errors} column counts the subset whose recorded outcome was an error.
\emph{Condense} counts admission-time shell-output reductions in the
\texttt{bash\_output\_condense} savings bucket. \emph{Decay} counts later
half-life events that render older tool results at smaller caps. Condensation
and decay are distinct stages. In the second table, \emph{checks} counts parsed
detector records. \emph{Valid} counts yes-or-no verdicts. \emph{Setup} counts
setup failures. \emph{Positive} counts hurdle-present verdicts. \emph{Applies}
and \emph{restores} count intervention overlay changes.

\begin{table}[H]
\centering
\caption{Observed context and command activity in the paired populations used for this analysis. Each entry after $n$ gives the event count followed by the number of sessions with at least one event; all listed markers are zero in the corresponding control sessions.}
\label{tab:channel-context-activity}
\scriptsize
\setlength{\tabcolsep}{3.5pt}
\begin{tabular}{lrrrrr}
\toprule
Cell & $n$ & command & errors & condense & decay \\
\midrule
Verified (20k) & 169 & 484/117 & 313/105 & 0/0 & 4079/140 \\
Verified (43k) & 169 & 786/150 & 435/142 & 0/0 & 3212/107 \\
Pro (49k) & 316 & 6926/236 & 6181/230 & 519/104 & 9318/246 \\
FeatureBench (47k) & 183 & 1250/131 & 1017/127 & 44/16 & 5057/143 \\
\bottomrule
\end{tabular}
\end{table}

\begin{table}[H]
\centering
\caption{Observed detector and conditioned-action activity in the same paired populations. Each entry gives the event count followed by the number of sessions with at least one event. The \emph{applies} count excludes later restorations; all listed markers are zero in the corresponding control sessions.}
\label{tab:channel-detector-activity}
\scriptsize
\setlength{\tabcolsep}{3.5pt}
\begin{tabular}{lrrrrrr}
\toprule
Cell & checks & valid & setup & positive & applies & restores \\
\midrule
Verified (20k) & 2193/168 & 2193/168 & 0/0 & 45/31 & 56/31 & 24/21 \\
Verified (43k) & 2895/169 & 2895/169 & 0/0 & 50/36 & 62/36 & 32/29 \\
Pro (49k) & 10109/314 & 10109/314 & 0/0 & 2464/153 & 633/153 & 240/145 \\
FeatureBench (47k) & 3490/181 & 3490/181 & 0/0 & 424/84 & 262/84 & 90/75 \\
\bottomrule
\end{tabular}
\end{table}

\FloatBarrier

\subsection{Selected context trajectories}

\begin{figure}[t]
\centering
\includegraphics[width=\linewidth]{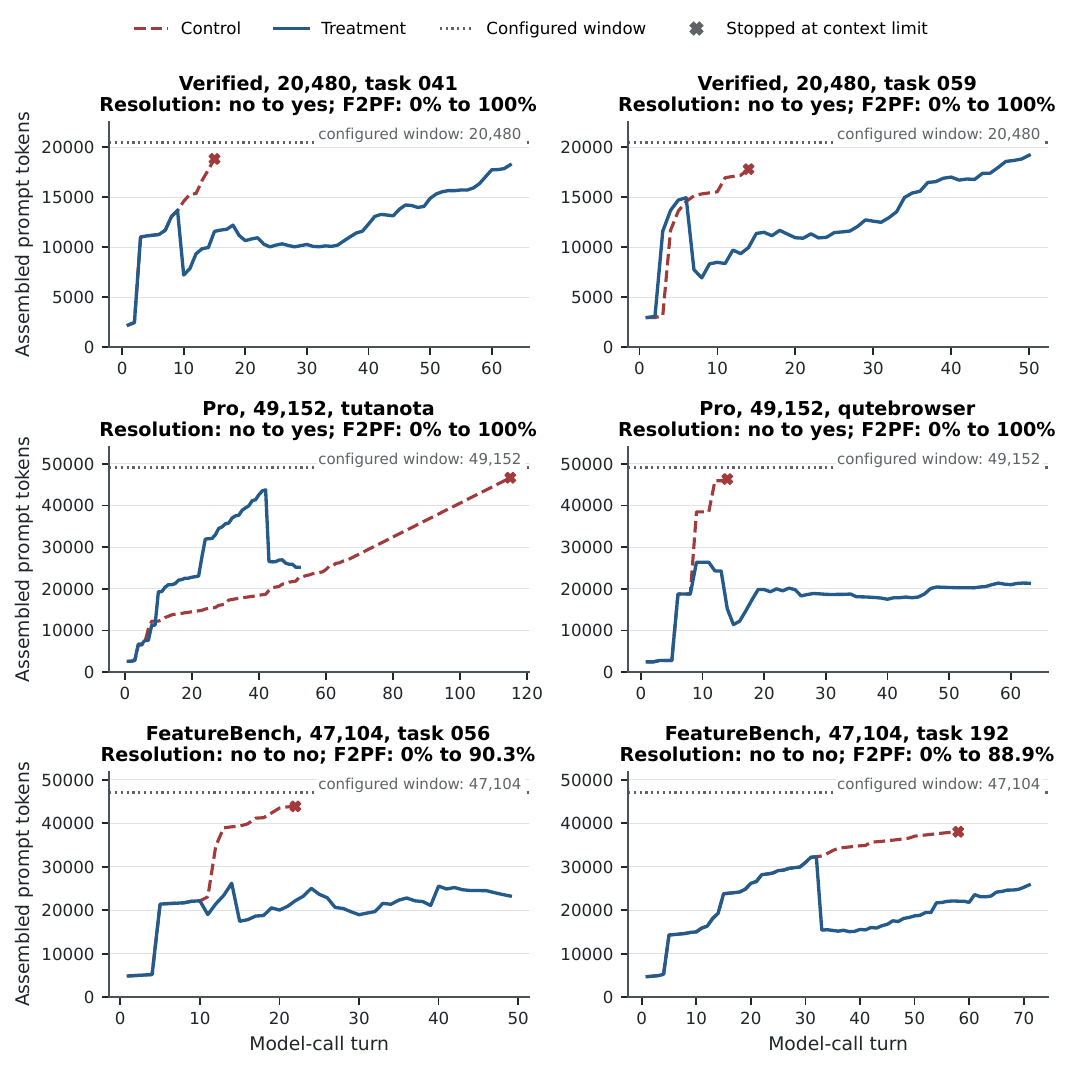}
\caption{Assembled prompt length by turn for six selected
task pairs, two per benchmark. The horizontal reference is the configured
window. Panel subtitles report control followed by treatment. The Verified
examples are treatment-only resolutions. The Pro and
FeatureBench examples have large positive paired F2PF differences.}
\label{fig:ctxcurves}
\end{figure}

Figure~\ref{fig:ctxcurves} records assembled prompt tokens at each model
call. A model output first enters this accounting when it becomes part of
the next prompt. When mechanical decay activates, older tool results are
rendered at smaller caps, and assembled length can drop. New tool results
enter at full size, so the input can grow again.

\FloatBarrier

\subsection{Exploratory context-policy probes}

Table~\ref{tab:dial-sweeps} compares the frozen settings with exploratory
context-depth and activation probes on \DialCohortN{} tasks per model. R6
scaled all Qwen3.6 age-tier caps, while R7/R8 varied activation and verbatim
protection for Qwen3.6 and R7 applied neighboring activation settings to
Devstral and Nemotron.

% AUTO-GENERATED by pipelines/make_result_tables.py
\begin{table}[H]
\centering
\scriptsize
\caption{Exploratory single-run probes of context depth and activation timing. Rows compare only with the frozen setting for the same model and cohort.}
\label{tab:dial-sweeps}
\begin{tabular}{lllrr}
\toprule
family & model & policy setting & solutions / $n$ & mean per-task F2PF (\%) \\
\midrule
R6 depth & Qwen3.6 & frozen caps, prior run & 6/15 & 45.3 \\
 &  & frozen caps, rerun & 7/15 & 47.5 \\
 &  & all caps $\times$0.5 & 5/15 & 45.0 \\
 &  & all caps $\times$2.0 & 4/15 & 33.3 \\
\addlinespace
R7/R8 timing & Qwen3.6 & activation 0.50, protect 4 & 3/15 & 22.2 \\
 &  & activation 0.25, protect 4 & 2/15 & 13.3 \\
 &  & activation 0.50, protect 2 & 2/15 & 15.6 \\
 &  & activation 0.25, protect 2 & 2/15 & 15.6 \\
 &  & activation 0.75, protect 4 & 2/15 & 13.3 \\
 &  & activation 0.50, protect 6 & 1/15 & 8.9 \\
 &  & activation 0.75, protect 6 & 2/15 & 15.6 \\
\addlinespace
R7 timing & Devstral & activation 0.50, protect 4 & 1/15 & 6.7 \\
 &  & activation 0.25, protect 4 & 1/15 & 10.0 \\
 &  & activation 0.50, protect 2 & 1/15 & 8.9 \\
 &  & activation 0.25, protect 2 & 2/15 & 15.6 \\
\addlinespace
R7 timing & Nemotron & activation 0.50, protect 4 & 1/15 & 6.7 \\
 &  & activation 0.25, protect 4 & 0/15 & 0.0 \\
 &  & activation 0.50, protect 2 & 0/15 & 0.0 \\
 &  & activation 0.25, protect 2 & 1/15 & 6.7 \\
\bottomrule
\end{tabular}
\end{table}

No altered setting improved resolution beyond the observed rerun difference.
Every Qwen3.6 variant was lower than its frozen baseline on both complete
solutions and mean per-task F2PF.

\subsection{Public artifacts}

The public repository at \PaperRepository{} contains the reusable Yuj harness.
The repository's
\href{https://github.com/sydches/yuj/tree/main/configs/paper}{\texttt{configs/paper/}}
directory records the ordered harness and serving layers for the four primary
Qwen3.6 pressure comparisons. Its
\href{https://github.com/sydches/yuj/tree/main/paper}{\texttt{paper/}}
directory contains the experimental contract, path-neutral cell provenance,
reading-analysis exclusion ledger, and machine-readable paired task outcomes
for the published outcome cells.

Benchmark task lists, evaluators, continuation launchers, scorer artifacts,
and raw runs are not part of that public repository.

\end{appendices}

\begingroup
\raggedright
\bibliographystyle{unsrt}
\bibliography{refs}
\endgroup
\end{document}